\documentclass{article}

\usepackage{arxiv}

\usepackage[utf8]{inputenc} 
\usepackage[T1]{fontenc}    
\usepackage{hyperref}       
\usepackage{url}            
\usepackage{booktabs}       
\usepackage{amsfonts}       
\usepackage{nicefrac}       
\usepackage{microtype}      
\usepackage{lipsum}		
\usepackage{graphicx}
\usepackage{natbib}
\usepackage{doi}
\usepackage{siunitx}

\title{WaveMix: Resource-efficient Token Mixing for Images}

\author{ \href{https://orcid.org/0000-0003-4110-9638}{\includegraphics[scale=0.06]{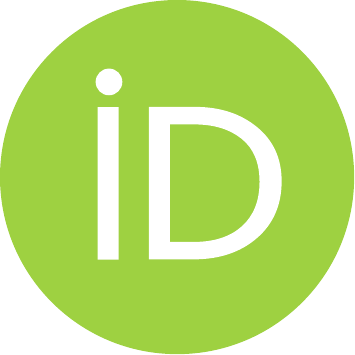}\hspace{1mm}Pranav Jeevan P} \\
	Department of Electrical Engineering\\
	Indian Institute of Technology Bombay\\
	Mumbai, India  \\
	\texttt{194070025@iitb.ac.in} \\
	\And
	\href{https://orcid.org/0000-0002-8003-6809}{\includegraphics[scale=0.06]{orcid.pdf}\hspace{1mm}Amit Sethi} \\
	Department of Electrical Engineering\\
	Indian Institute of Technology Bombay\\
	Mumbai, India \\
	\texttt{asethi@iitb.ac.in} \\
}

\date{}

\renewcommand{\headeright}{Preprint}
\renewcommand{\undertitle}{Preprint}
\renewcommand{\shorttitle}{WaveMix}

\hypersetup{
pdftitle={Wav},
pdfsubject={q-bio.NC, q-bio.QM},
pdfauthor={Pranav Jeevan, Amit Sethi},
pdfkeywords={Transformer, Image classification, Linear Attention, Deep Learning},
}

\begin{document}
\maketitle

\begin{abstract}
Although certain vision transformer (ViT) and CNN architectures generalize well on vision tasks, it is often impractical to use them on green, edge, or desktop computing due to their computational requirements for training and even testing. We present WaveMix as an alternative neural architecture that uses a multi-scale 2D discrete wavelet transform (DWT) for spatial token mixing. Unlike ViTs, WaveMix neither unrolls the image nor requires self-attention of quadratic complexity. Additionally, DWT introduces another inductive bias -- besides convolutional filtering -- to utilize the 2D structure of an image to improve generalization. The multi-scale nature of the DWT also reduces the requirement for a deeper architecture compared to the CNNs, as the latter relies on pooling for partial spatial mixing. WaveMix models show generalization that is competitive with ViTs, CNNs, and token mixers on several datasets while requiring lower GPU RAM (training and testing), number of computations, and storage. WaveMix have achieved State-of-the-art (SOTA) results in EMNIST Byclass and EMNIST Balanced datasets. 

\end{abstract}


\section{Introduction}
\label{sec:intro}
Most of the neural architectures that generalize well (infer accurately) on vision applications require power-hungry and expensive GPUs to train in a reasonable time, which is a bottleneck for many practical applications. This is especially true of vision transformer (ViT) architectures due to their use of self-attention~\cite{dosovitskiy2021image}, but is also a concern for convolutional neural networks (CNNs). Our objective was to propose and test neural architectures for image classification that use significantly fewer computational resources (GPU RAM for a fixed batch size, images per second, and model storage size) for training and testing, and yet generalize competitively with the state-of-the-art ViTs and CNNs.

The self-attention mechanism in the transformers~\cite{vaswani2017attention} model long-range relationships between tokens and gives state-of-the-art generalization in NLP and image recognition~\cite{dosovitskiy2021image}. However, the quadratic complexity of self-attention with respect to the sequence length (number of pixels in an unrolled image) creates a computational challenge for training ViT models. To some extent, this challenge has been alleviated by the use of sparse and linear attention mechanisms~\cite{Tay2020EfficientTA}. However, we believe that the computational burden can be further reduced by using an appropriate inductive bias for images, which the transformers lack.

On the other hand, convolutional neural networks (CNNs) have the inductive bias to handle 2D images, such as the translational equivariance of convolutional layers and partial scale invariance of the pooling layers, which enables them to generalize well with smaller image datasets, while also consuming fewer computational resources. However, CNN layers are not well-structured to capture the long-range dependencies compared to self-attention models due to the local scope of convolutional and pooling operations. As a result, CNNs require more layers to increase their receptive fields (spatial token mixing) compared to ViTs.

More recently, hybrid vision X-formers~\cite{Jeevan_2022_WACV} that combine inductive priors of convolutional layers with relatively efficient long-range token mixing of linear attention mechanisms have been proposed~\cite{choromanski2021rethinking,xiong2021nystromformer}. However, their data and computational requirements still remain impractical for many applications. By proposing WaveMix\footnote{Our code is available at \url{https://github.com/pranavphoenix/WaveMix}}, we take a step in the search for novel hybrid architectures for vision and further reduce the data and computational requirements for generalization comparable to ViTs and CNNs.

We replace the learnable attention mechanism in hybrid transformers with predefined (unlearnable) multi-level DWT layers for spatial token mixing. Our motivation is to utilize the well-researched multi-scale analysis properties of wavelet decomposition for image processing~\cite{Kingsbury97imageprocessing}. However, unlike previous works we do not simply use wavelet transforms to extract image features that are passed to a machine learning model. Instead, our architecture starts with a convolutional layer for short-range feature extraction (image-specific inductive bias), and then alternate between multi-level DWT for long-range spatial token mixing and convolutional layers for channel (and some spatial) token mixing. We thus obviate the need for image unrolling and even linear attention mechanisms. Additionally, wavelet decomposition introduces another form of image-specific inductive bias, that is hitherto unused in popular neural network architectures.

WaveMix achieves state-of-the-art (SOTA) generalization on EMNIST Balanced and Byclass datasets and performs better than transformers, ResNets and other token mixers in all the other datasets. It consumes orders of magnitude less GPU RAM than transformer models. When compared to CNNs, WaveMix performs on par with the deeper ResNets while using fewer parameters, layers, and GPU RAM. All of our experiments were done on a single GPU with 16GB RAM.

\section{Related Works}

\textbf{Token mixing for images:} Experiments have shown that replacing the self-attention in transformers with fixed token mixing mechanisms, such as the Fourier transform (FNet), achieves comparable generalization with lower computational requirements~\cite{leethorp2021fnet}. Other token-mixing architectures have also been proposed that use standard neural components, such as convolutional layers and multi-layer perceptrons (MLPs) for mixing visual tokens. MLP-mixer~\cite{tolstikhin2021mlpmixer} uses two MLP layers (cascade of 1x1 convolutions) applied first to image patch sequence and then to the channel dimension to mix tokens. ConvMixer~\cite{trockman2022patches} uses standard convolutions along image dimensions and depth-wise convolutions across channels to mix token information. These token mixing models perform well with lower computational costs compared to transformers without compromising generalization. The quadratic complexity with respect to the sequence length (number of pixels) for vanilla transformers has also led to the search for other linear transforms to efficiently mix tokens~\cite{Jeevan_2022_WACV}. 

\textbf{Wavelets for images:} Extensive prior research has uncovered and exploited various multi-resolution analysis properties of wavelet transforms on image processing applications, including denoising~\cite{5598411}, super-resolution~\cite{8014882}, recognition~\cite{8536280}, and compression~\cite{136601}. Features extracted using wavelet transforms have also been used extensively with machine learning models~\cite{1028240}, such as support vector machines and neural networks~\cite{7754470}, especially for image classification~\cite{Nayak2016BrainMI}. Instances of integration with neural architectures include the following. ScatNet architecture cascades wavelet transform layers with non-linear modulus and average pooling to extract a translation invariant feature that is robust to deformations and preserves high-frequency information for image classification~\cite{6522407}. WaveCNets replaces max-pooling, strided-convolution, and average-pooling of CNNs with DWT for noise-robust image classification~\cite{li2020wavelet}. Multi-level wavelet CNN (MWCNN) has been used for image restoration as well with U-Net architectures for better trade-off between receptive field size and computational efficiency~\cite{liu2018multilevel}. Wavelet transform has also been combined with a fully convolutional neural network for image super resolution ~\cite{kumar2017convolutional}.

We propose using the two-dimensional discrete wavelet transform (2D DWT) for long-range token mixing. Among the different types of mother wavelets available, we used the Haar wavelet (a special case of the Daubechies wavelet~\cite{57199}) also known as Db1), which is frequently used due to its simplicity and faster computation. Haar wavelet is both orthogonal and symmetric in nature, and has been used to extract basic structural information from images~\cite{Porwik2004TheHT}.

\section{WaveMix Architecture}

\begin{figure}
  \centering
   \includegraphics[width=0.7\linewidth]{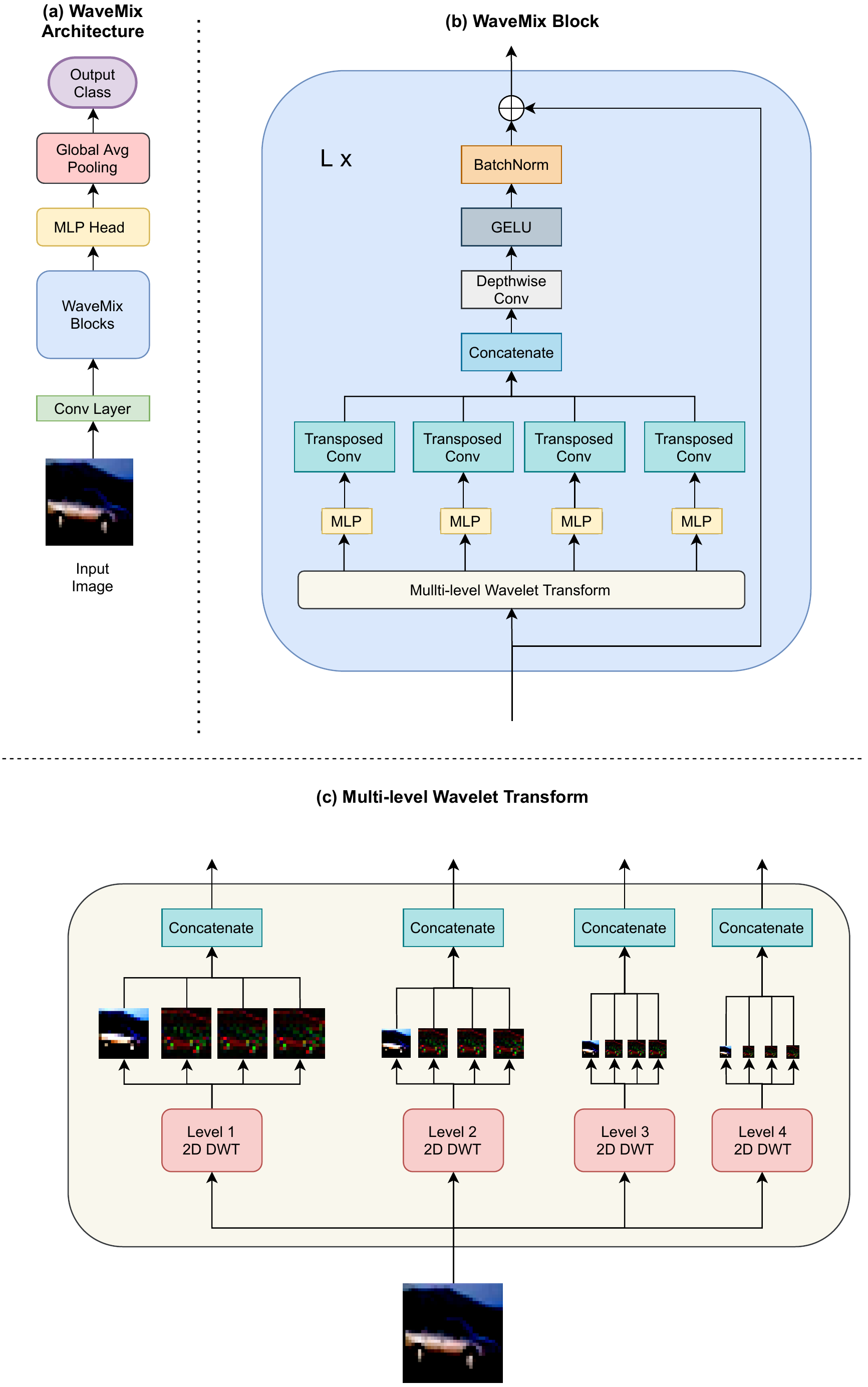}

   \caption{WaveMix Architecture: (a) Overall architecture with initial convolutional layer, WaveMix blocks, global average pooling, and the final classification head; (b) details of the WaveMix block used in the overall architecture; and (c) representation of the multiple levels of the wavelet transform used in a WaveMix block.}
   \label{fig:wave}
\end{figure}

Image pixels have several interesting co-dependencies. The localized and stationary nature of certain image features (e.g., edges) has been exploited using linear space-invariant filters (convolutional kernels) of limited size. Scale invariance of natural images has been exploited to some extent by pooling~\cite{lecun-gradientbased-learning-applied-1998}. However, we think that scale invariance can be better modeled by wavelet decomposition due to its natural multi-resolution analysis properties. Additionally, the finer scale of a multi-level wavelet decomposition also incorporates the idea of linear space-invariant feature extraction using convolutional filters of small support; albeit it uses predefined weights. The basic idea, therefore, behind our proposed architecture is to alternate between spatially repeated (convolutional) learnable feature extraction and fixed multi-resolution token mixing using DWT for a few layers. Injecting learnability is key to improving the utility of the wavelet transform, while convolutional kernels allow parameter-efficient learning suitable for the location-invariant statistics of images. This combination requires far fewer layers and parameters than using only convolutional layers with pooling. On the other hand, while transformers and other token mixers have very large effective receptive fields right from the first few layers, they do not utilize inductive priors that are suitable for images. This is where the wavelet transform plays its role.

\subsection{Overall Architecture}

As shown in Fig. \ref{fig:wave}(a), in our models the input image is first passed through a convolutional layer that creates feature maps of the image. The use of trainable convolutions \emph{before} the wavelet transform is a key aspect of our architecture, as it allows the extraction of only those feature maps that are suitable for the chosen wavelet family. This is followed by a series of WaveMix blocks, an MLP head, and global average pooling layer, and an output layer for classification. The global average pooling imparts a certain degree of size invariance to our architecture.



At no point in the model do we unroll the image into a sequence of pixels. So, we have developed a model that can exchange information between pixels which are separated by long distances without using self-attention, thereby escaping the quadratic complexity bottleneck of self-attention. WaveMix  even eliminates the learning mechanism required for linear approximations to the quadratic attention.

\subsection{WaveMix Block}

As shown in Fig. \ref{fig:wave}(b), inside the WaveMix block, the input channels are decomposed into multiple levels of 2D DWT, which produces four output channels (one approximation and three details) for each input channel per DWT level. Channel mixing and reduction is performed by the MLP (two 1x1 convolutional layers separated by a GELU nonlinearity). Channel size reconciliation between multiple levels of DWT is performed using transposed convolutions (up-convolutions). The kernel size and stride of deconvolutional layers were chosen such that all the different sized outputs from different levels of DWT were brought back to the same size as the original image. We chose deconvolutional layer rather than an inverse DWT because the former is much faster and consumes less GPU than the latter. The outputs from the deconvolutional layers are then concatenated together (depth or channel-wise) and this output has the same number of channels as the input to the WaveMix block (embedding dimension). The concatenated output is then passed through a depth-wise convolutional layer with a kernel size of $5$, followed by GELU activation and batch normalization. A residual connection~\cite{he2015deep} is provided within each WaveMix block so that the model can be made deeper with a larger number of blocks, if necessary. 

The approximation and detail coefficients are extracted from the input using multi-level 2D DWT, as shown in Fig.~\ref{fig:wave} (c). We used Haar wavelet (Db1) for generating the 2D DWT output \footnote{Base code: \url{https://pytorch-wavelets.readthedocs.io/en/latest/readme.html}}. In addition to the simplicity of implementation, an additional advantage of the Haar wavelet is that it reduces the size of a feature map exactly by a factor of 2$\times$2, which makes the design of the deconvolution layer to increase the size back to the original easier. The number of levels of wavelet decomposition needed is decided based on the image size. Each level reduces the DWT output size by half. Therefore, we use as many levels as necessary till the input size is reduced to 2$\times$2 to ensure token mixing over long spatial distances. For example, a 32$\times$32 image requires a 4-level 2D DWT, which creates 16$\times$16, 8$\times$8, 4$\times$4 and 2$\times$2 sized outputs, respectively at different levels. To the the low-resolution image generated at each level (approximation sub-band) we concatenate in the channel dimension the corresponding three sets of detail coefficients from the same level. Hence, each level will output different sized images (16$\times$16, 8$\times$8, 4$\times$4 and 2$\times$2) each having 4 times more feature maps than the input embedding dimension.




We conjecture that the lower levels of the DWT capture the finer details of the image while higher levels capture more global information. The feed-forward (MLP or 1x1 convolutional) sub-layers immediately following this DWT have access only to the outputs at the corresponding level to learn the features. Once the features learned from each resolution level are passed through transposed convolutions, where all the different low-resolution images are up-sampled to full image size and concatenated along channel dimension, the succeeding depth-wise convolutional layer will have full access to all the local and global information carried by the tokens. Transposed convolution of the lower resolution DWT outputs will spread the global information to all regions of the image which helps the succeeding sub-layers model relationships between tokens both locally and globally. 

The depth-wise convolutional block processes the combined concatenated feature maps containing information in multiple resolutions of the image. This enables the model to mix information from different resolutions of the image along with mixing of global information from different spatial locations. 

The presence of normalization and residual connections enable the construction of deeper models that can handle larger images.

\subsection{WaveMix-lite Block}

\begin{figure}
  \centering
   \includegraphics[width=0.45\linewidth]{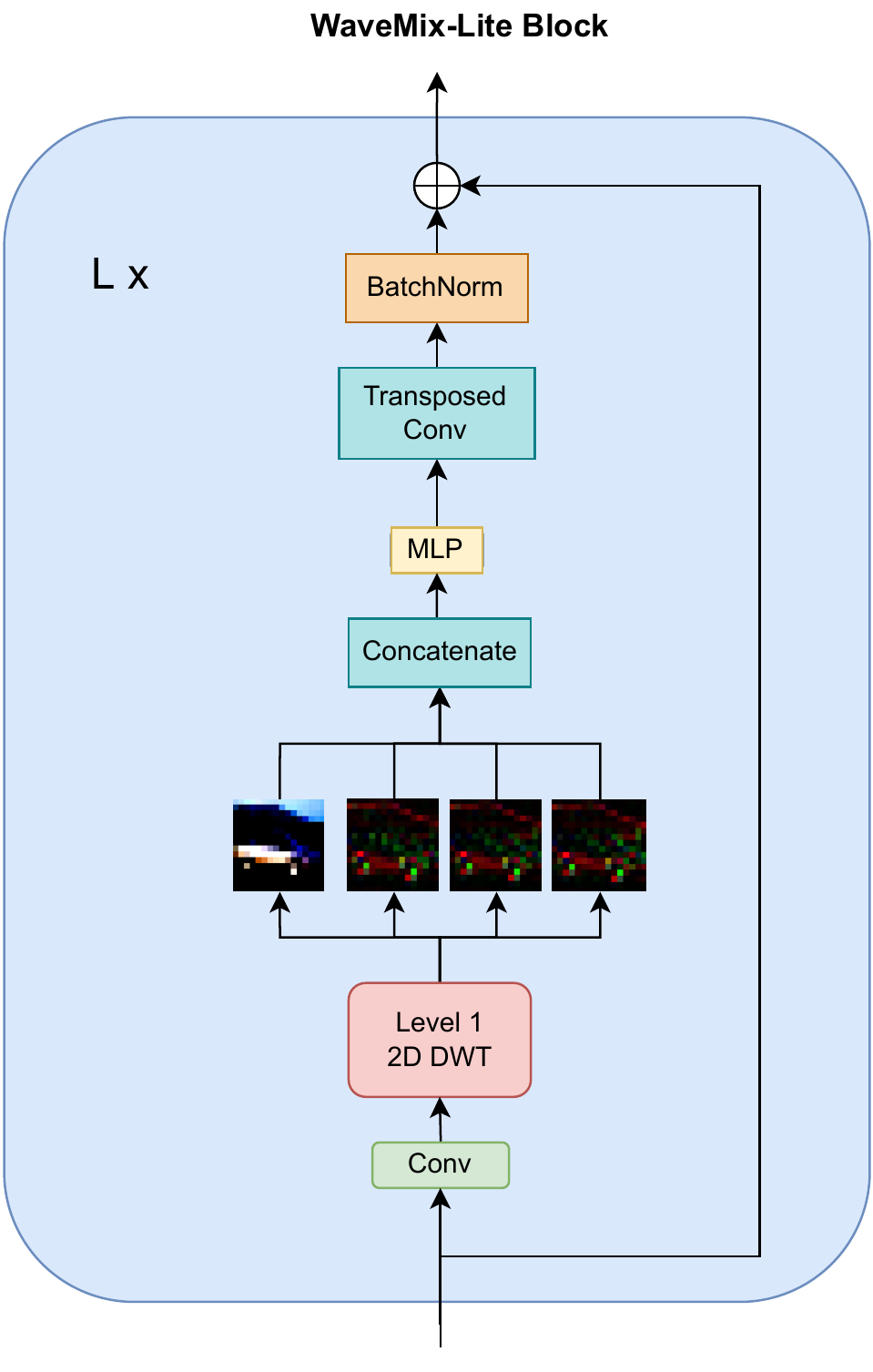}
   \caption{WaveMix-Lite Block}
   \label{fig:wavelite}
\end{figure}

We created a lighter and faster version of WaveMix block as shown in Fig.~\ref{fig:wavelite} to fit models with larger embedding dimensions into a single GPU with 16 GB RAM. First, the input to the WaveMix-Lite is passed through a convolutional layer which decreases the embedding dimension by four, so that the concatenated output after 2D DWT has the same dimension as input. To reduce the parameters and computations, we only use 1 level 2D DWT in WaveMix-Lite. The concatenated output from 2D DWT is passed to an MLP layer with GELU non-linearity having a multiplication factor more than one. This output is then passed through resizing deconvolutional sub-layer and then through a batch normalization sub-layer. A residual connection~\cite{he2015deep} is also provided within each WaveMix-Lite block. We remove the GELU non-linearity and depth-wise convolutional sub-layers used at the end of WaveMix block. These changes significantly reduce the number of parameters and GPU footprint of the model and increase its training speed, which are extremely useful while training large datasets. WaveMix-Lite is mostly used when we need high embedding dimensions. The larger embedding dimension available will ensure that all information about long-range global dependencies are passed to subsequent layers in one level of 2D DWT, without the need for multiple levels. We replace the WaveMix block with WaveMix-Lite block when using models with embedding dimensions larger than 64.

\section{Experimental Settings}

\subsection{Datasets and models compared}

To demonstrate the general applicability of WaveMix, we used multiple types of image datasets based on the number of images and image size. Small datasets of smaller image sizes included CIFAR-10, CIFAR-100~\cite{Krizhevsky09learningmultiple}, EMNIST~\cite{7966217}, Fashion MNIST~\cite{DBLP:journals/corr/abs-1708-07747}, and SVHN~\cite{Netzer2011}. Small datasets of larger image sizes included STL-10~\cite{pmlr-v15-coates11a}, Caltech-256~\cite{griffin2007caltech} and Tiny ImageNet~\cite{Le2015TinyIV}. We also used larger datasets with reduced image size (e.g., 64x64), such as Places-365~\cite{zhou2017places}, ImageNet-1k~\cite{5206848}, and iNaturalist 2021-10k (iNAT mini)~\cite{horn2021benchmarking}. For comparison, we chose ResNet-18 and ResNet-34, ResNet-50 and ResNet-101~\cite{he2015deep} as convolutional models; FNet~\cite{leethorp2021fnet}, MLP-Mixer~\cite{tolstikhin2021mlpmixer}, and ConvMixer~\cite{trockman2022patches} as token-mixers; and ViT~\cite{dosovitskiy2021image}, Hybrid ViN~\cite{Jeevan_2022_WACV} CCT\footnote{Base code: \url{https://github.com/lucidrains/vit-pytorch}}~\cite{hassani2021escaping}, and CvT~\cite{wu2021cvt} as transformer models. The 2D versions of FNet and MLP-Mixer were also used for the experiments, as explained next.

\subsection{Modifications made to 1D Token Mixers}

Recently, it has been shown that 1-D Fourier transform of FNet and 1-D MLP-Mixer can be used for image classification without much modifications to give better results than a ViT of a comparable size~\cite{Jeevan_2022_WACV}. However, in both those models the image had to be unrolled as a sequence of patches. We wanted to see if using a 2-D version of these models, where image can be processed in its 2-D form, can provide an advantage. That is, we compared with token mixing architectures that used a linear transform without learnable parameters (FNet) or a nonlinear transform with learnable parameters (MLP-Mixer) to understand if the wavelet transform is better suited for images than these other transforms. For a comparison of the performance of WaveMix architecture with the appropriate alternatives, we re-designed other 1D-token mixing architectures to suit image data. Since ConvMixer architecture was designed to handle image data in 2D form, no modifications were made to it.

The FNet~\cite{leethorp2021fnet} architecture was built to handle long 1D sequence and not 2D images. So the model used by~\cite{Jeevan_2022_WACV} in their experiments, which unrolled the image into pixel sequences, did not perform well. We modified the 1D FNet by replacing 1D Fourier transform with a 2D one, and then applied a 1D Fourier transform across the channel dimension and took the real parts. This enabled the information to mix across all the three dimensions of the image. The feed-forward layers were implemented using 1$\times$1 kernel convolutional layers along with 2D batch normalisation. A residual connection~\cite{he2015deep} was added across the 2D FNet block. This 2D FNet block replaces the WaveMix block in the WaveMix architecture in our experiments.

The MLP-Mixer~\cite{tolstikhin2021mlpmixer} was also redesigned to handle 2D image data by applying two MLPs along both height and width dimensions and another MLP along the channel dimension. We used layer normalization before the MLPs acting along height and width dimensions and a 2D batch normalization before the MLP along channel dimension. The three MLPs enable the mixing of tokens along the three dimensions of an image. A residual connection~\cite{he2015deep} was added across the 2D MLP-Mixer block. This 2D MLP-Mixer block replaces the WaveMix block in the WaveMix architecture in our experiments.

\subsection{Training and Architectural Details}

We trained models using Adam optimizer ($\alpha = 0.001, \beta_{1} = 0.9, \beta_{2}=0.999, \epsilon = 10^{-8}$) with a weight decay coefficient of 0.01. We used automatic mixed precision in PyTorch during training to optimize speed and memory consumption. Experiments were done with a 16 GB Tesla V100-SXM2 GPU available in Google Colab Pro Plus. No image augmentations were used while training the models. GPU usage for a batch size of 64 was reported along with top-1\% accuracy from best of three runs with random initialization based on prevailing protocols~\cite{hassani2021escaping}. Maximum number of epochs in all experiments was set as 120.


Patch size of 1 was chosen for all the transformer models that unrolled the images as a sequence of pixels, such as the ViT. ConvMixer with kernel size of 8 and patch size of 1 was used for $32 \times 32$ images and patch size of 2 for $64 \times 64$ images. We used 64 benchmark points in the Nyströmformer used in hybrid ViN. A dropout of 0.5 was used across all models.

In WaveMix, we applied two layers of 3$\times$3 convolutions to the input image. These layers increased the channel dimension from three to the set embedding dimension in two stages. We observed that since we are constrained to use a single 16 GB GPU, we could not send image of size larger than $64 \times 64$ into the WaveMix-Lite block for a batch size of 64. For images with resolutions larger than $64 \times 64$, we adjusted the stride so that the output from the convolutional layers reduced the image resolution to $64 \times 64$. Stride of 2 was used in both layers when we input $256 \times 256$ images. For images of size less than $64 \times 64$, we set the stride to 1. Unless otherwise stated, all WaveMix models with embedding dimension of 128 and 256 used WaveMix-Lite blocks.

\section{Results}

\begin{table}[]
\centering
\caption{Comparison of top-1 accuracy (without data augmentation) and computational requirements of various models for image classification}
\begin{tabular}{@{}lccccc@{}}
\hline
\multicolumn{1}{l}{Models} &
  \begin{tabular}[c]{@{}c@{}}\#Param\\ (Million)\end{tabular} &
  \begin{tabular}[c]{@{}c@{}}GPU\\ (GB)\end{tabular} &
  \begin{tabular}[c]{@{}c@{}}CIFAR -10\\ acc. (\%)\end{tabular} &
  \begin{tabular}[c]{@{}c@{}}CIFAR-100\\ acc. (\%)\end{tabular} &
  \begin{tabular}[c]{@{}c@{}}Tiny ImageNet\\ acc. (\%)\end{tabular} \\ \hline
\multicolumn{6}{l}{\textit{Convolutional Models}}  
\\ \hline
ResNet-18                 & 11.2 & 1.2          & 86.29 & 59.15               & 48.11                \\
ResNet-34                 & 21.3 & 1.4          & 87.97 & 57.79                & 45.60                \\
ResNet-50                 & 25.2 & 3.3          & 86.21 & 58.48                & 48.77                \\ \hline
\multicolumn{6}{l}{\textit{Token Mixers}}                                                             \\ \hline
2D FNet-64/5              & 0.10 & 1.0          & 64.60         &   29.32            &        15.53              \\
2D FNet-128/5             & 0.41 & 1.6          & 70.52          &       32.13       &                26.56      \\
2D MLP-Mixer-64/5         & 0.15 & 1.8          & 47.81          &   20.76                   &      7.78                \\
2D MLP-Mixer-128/5        & 0.45 & 3.7          & 55.02    & 22.81      &  11.12\\
ConvMixer-256/8           & 0.67 & 3.7           & 85.41        &     57.29            &        43.15              \\
ConvMixer-256/16          & 1.3  & 7.0          & 88.46 & 61.80                & 45.39                \\ \hline
\multicolumn{6}{l}{\textit{Transformer Models}}                                                       \\ \hline
ViT-$128/4\times4$        & 0.53 & 13.8         & 56.81          &            30.25          & 26.43                \\
Hybrid ViN-$128/4\times4$ & 0.62 & 4.8          & 75.26          &    51.44                  & 34.05                \\
CCT-$128/4\times4$        & 0.90 & 15.8         & 82.23          &   57.09           & 39.05                \\
CvT-$128/4\times4$        & 1.10 & 15.4         & 79.93          &       48.29            & 40.69                \\ \hline
\multicolumn{6}{l}{\textit{WaveMix Models}}                                                           \\ \hline
WaveMix-16/5              & 0.18 & 0.2 & 78.04 &   34.32           &       26.96               \\ 
WaveMix-32/5              & 0.72 & 0.2 & 81.47 &   45.70              &    29.97                  \\
WaveMix-64/5              & 2.88 & 0.3 & 86.16 &  56.20              &  38.19\\ 
WaveMix-128/7              & 2.42 & 1.3 & \textbf{91.08} &   68.40          &       \textbf{52.03}               \\ 
WaveMix-256/7              & 9.62 & 2.3 & 90.72 &      \textbf{70.20}        &    51.37                 \\ 
\hline
\end{tabular}
\label{tab:perf}
\end{table}

\begin{table}[]
\centering
\caption{Top-1 accuracy of WaveMix compared to ResNets on different EMNIST datasets}
\label{tab:emnist}
\begin{tabular}{@{}lcccccccc@{}}
\hline
Models    & \#Params & Byclass & Bymerge & Letters & Digits & Balanced & MNIST & Fashion \\ \hline
ResNet-18 & 11.2   & 87.98   & 91.09   & 94.76   & 99.67  & 89.00    & 99.69 & 93.35   \\
ResNet-34 & 21.3   & 88.10    & 91.13   & 95.04   & 99.68  & 89.17    & 99.67 & 93.34   \\
ResNet-50 & 23.6   & 88.18   & 91.29   & 94.64   & 99.62  & 89.76    & 99.56 & 93.30   \\ \hline
WaveMix-128/7 & \textbf{2.4}   &  \textbf{88.43}  &  91.52  &  \textbf{95.78} & \textbf{99.77}  &  \textbf{91.06}   & \textbf{99.71} &  \textbf{93.91}\\
WaveMix-256/7 & 9.6 & 88.42 & \textbf{91.59} & 95.56 & 99.70 & 90.36 & 99.65 & 93.78 \\ \hline
\end{tabular}
\end{table}

\begin{table}[]
\centering
\caption{Top-1 accuracy of WaveMix compared to ResNets on datasets of different image resolutions}
\label{tab:resolution}
\begin{tabular}{@{}lcccccc@{}}
\hline
\multicolumn{1}{l}{Models} &
  \begin{tabular}[c]{@{}c@{}}STL-10\\ $96 \times 96$\end{tabular} &
  \begin{tabular}[c]{@{}c@{}}SVHN\\ $32 \times 32$\end{tabular} &
  \begin{tabular}[c]{@{}c@{}}Caltech-256\\ $256 \times 256$\end{tabular} &
  \begin{tabular}[c]{@{}c@{}}Places-365\\ $256 \times 256$\end{tabular} &
  \begin{tabular}[c]{@{}c@{}}iNAT-2021\\ $256 \times 256$\end{tabular}\\ \hline
ResNet-18          & 70.41          & 97.40          & 52.97     & 48.74  & 26.35   \\
ResNet-34            & 68.07          & 97.47          & 50.92   & 49.02  & 31.02      \\
ResNet-50            & 66.04          &   97.32             & 49.97    & 49.80 &  33.14    \\ \hline
WaveMix-256/7  & \textbf{70.88} & \textbf{97.61} & \textbf{54.62} & \textbf{49.83} & \textbf{33.23} \\ \hline
\end{tabular}
\end{table}
\setlength{\tabcolsep}{2 pt}

\subsection{Notation}

We use the format \emph{Model Name-Embedding Dimension/Layers $\times$ Heads} for transformer based models and same notation without \emph{heads} for other architectures. For example, CCT with embedding dimension of 128 having 4 layers and 4 heads is labelled as CCT-$128/4\times4$.

\subsection{Main Results}

Table~\ref{tab:perf} shows the performance of WaveMix compared to other architectures on image classification using supervised learning on different datasets. WaveMix models outperform ResNets, transformers, hybrid xformers, and other token-mixing models while requiring the least GPU RAM for the same batch size. WaveMix-128 and WaveMix-256 achieve superior generalization, while WaveMix-64/5 achieves generalization similar to ResNets and ConvMixer with only 21\% of GPU RAM used by ResNets and 5\% to 10\% used by ConvMixer. WaveMix-256 needs only 70\% of the GPU RAM used by ResNet-50. Even the smallest WaveMix-16/5 model outperforms the transformer models (ViT and hybrid ViN) by significant margins while consuming less than 2\% of GPU RAM compared to ViT and 6\% of that compared to hybrid ViN. 

ConvMixer performed better than FNet and MLP-Mixer because, just like wavelet transforms, convolutions also possess an inductive prior for exploiting spatial invariance in 2D image data. However, the higher accuracy obtained by WaveMix is due its ability to process an image at multiple resolutions in parallel, where it can learn image features obtained at different scales. This ability is absent in convolutional layers, which require pooling for large-scale information mixing, followed by quadratic attention or its linear approximations. The low GPU usage of WaveMix is due to the linearity of wavelet transforms which is easy to compute compared to convolutions and expensive self-attention matrices. The low GPU consumption of WaveMix is noteworthy when we add the fact that our model computes not just one, but four stages of wavelet transforms and processes them in parallel to get the output.

The 2D FNet and 2D MLP-Mixer that use Fourier transforms and MLP, respectively, for token-mixing, could not match the generalization of WaveMix. This is due the ability of wavelet transform to better handle  multi-resolution token-mixing for images, which is absent in these other two models. Although the Fourier transform is also in some sense a multi-resolution transform, it suffers from non-local analysis even for fine details, which is precisely why wavelets and cosine transform had replaced it for various image analysis tasks, such as image compression. This comparison with FNet and MLP-Mixer confirms that the presence of wavelet transform in our architecture is essential for the improved accuracy we observe in our models, since the other network components, such as feed-forward layers are present in all three token-mixing models.

\subsection{In-depth Comparison with ResNets}

Even though convolution has been widely regarded as a GPU-efficient operation, the need for deeper layers have necessitated the use of networks having over tens to hundreds of layers for achieving high generalization. Even though a single convolutional operation is comparatively cheaper than a 2D DWT, we can achieve generalization comparable to deep convolutional networks with very few layers of wavelet transforms. This ability of wavelet transform to provide competitive performance without needing large number of layers helps in improving the efficiency of the network by consuming much lesser GPU RAM than deep convolutional models like ResNets. We can see from Table~\ref{tab:emnist} that WaveMix models outperforms ResNets on all the EMNIST datasets ($28 \times 28$).  WaveMix achieved state-of-the-art (SOTA) results in EMNIST Balanced dataset (0.01 percentage points more than VGG-5 network with SpinalNet classification layers~\cite{Kabir2020SpinalNetDN}) and EMNIST Byclass dataset (0.31 percentage points more than the previous best~\cite{7966217}).


Since the 2D DWT is a linear transformation, no learnable parameters are needed for token mixing in WaveMix. Hence, we observe that WaveMix requires significantly fewer parameters than ResNets.

We see from Table~\ref{tab:resolution} that even when we test on larger datasets with millions of images and datasets having image sizes greater than $64 \times 64$, the performance of the WaveMix model is still better than the ResNets. Datasets having hundreds of million images pose real challenges to the hardware available to most researchers. Even though image resolution had to be downscaled to $64 \times 64$ using the initial convolutional layers, it does not affect the performance of the WaveMix models and they perform competitively with ResNets.

\begin{table}[]
\centering
\caption{Generalization and GPU RAM usage for a batch size of 64 and maximum batch size possible in one 16 GB GPU for WaveMix and ResNets on ImageNet-1k dataset with image size downscaled to $64 \times 64$}
\begin{tabular}{@{}lcccc@{}}
\hline
Models & GPU(GB) & Top-1 Acc. (\%) & Top-5 Acc. (\%) & Max Batch Size \\ \hline
ResNet-18    & 2.6 & 50.67              & 75.07 & 384\\   
ResNet-34     & 3.5 & 55.04         & 76.95 & 288\\   
ResNet-50     & 11.3& 55.66              & 78.40 & 96\\   
ResNet-101    &  15.1& 56.05         & 79.43  & 64\\   \hline
WaveMix-256/7  & 9.0  & \textbf{56.66}   & \textbf{80.04} & 112 \\ \hline
\end{tabular}
\label{tab:imagenet}
\end{table}

We have also experimented with ImageNet-1k, although the image size was downscaled to $64 \times 64$ due to computational and storage constraints. We see from Table~\ref{tab:imagenet} that WaveMix can outperform the deep ResNet models like ResNet-50 and ResNet-101 while using less GPU RAM. 

The results from training large datasets including ImageNet-1k, iNaturalist-10k, and Places-365 shows that WaveMix is resource-efficient and performs on par with the ResNets even for datasets with large image sizes. Thus, WaveMix can allow practitioners to pre-train models with large datasets with millions of images using less GPUs, thus opening more possibilities for their applications.

\subsection{Training and Inference Speed}

\begin{table}[]
\centering
\caption{Training (one forward and backward passes) and inference speeds (images/s) of various models for images sizes of $32 \times 32$ and $64 \times 64$ on a 16GB GPU}
\begin{tabular}{@{}lcccc@{}}
\hline
\multicolumn{1}{l}{Models} &
  \begin{tabular}[c]{@{}c@{}}Train\\ $32 \times 32$\end{tabular} &
  \begin{tabular}[c]{@{}c@{}}Infer\\ $32 \times 32$\end{tabular} &
  \begin{tabular}[c]{@{}c@{}}Train\\ $64 \times 64$\end{tabular} &
  \begin{tabular}[c]{@{}c@{}}Infer\\ $64 \times 64$\end{tabular} \\ \hline

ResNet-18                 & 1571          & 3436          & 467          & 1389          \\
ConvMixer-256/16    & 166           & 451           & 120          & 445           \\
Hyb. ViN-128/$4 \times 4$ & 243           & 801           & 240          & 773           \\
CCT-128/$4 \times 4$        & 149           & 521           & 149          & 514           \\
\hline
WaveMix-16/5              & \textbf{3230} & \textbf{4310} & \textbf{797} & \textbf{2208} \\
WaveMix-32/5              & 1834          & 3802          & 460          & 1623          \\
WaveMix-128/7              & 1279           & 3401           &   276        &    488        \\ 
WaveMix-256/7              &   495         &   1028       &     159     &   408       \\ \hline
\end{tabular}
\label{tab:speed}
\end{table}

Table~\ref{tab:speed} shows that WaveMix is significantly faster in training and inference than ConvMixer and transformers as it does not have the complexity of self-attention. WaveMix's speed is comparable to shallower ResNets, which can be attributed to WaveMix's ability to learn useful image representations with just few layers compared to CNNs. The lack of significant differences in training and inference speeds of ConvMixer and transformer models between the 2 image sizes is due to the variation in patch sizes and strides which essentially reshapes the $64\times64$ image to $32\times32$.

\begin{table}[]
\centering
\caption{Maximum train and test batch size possible for various models on a 16 GB GPU for training on CIFAR-100 datset}
\label{tab:bs}
\begin{tabular}{@{}lcc@{}}
\hline
\multicolumn{1}{l}{Model} & Max batch size for training & Max batch size for inference\\ \hline
ResNet-50                 & 320                  & 960                \\ \hline
WaveMix-128               & \textbf{768}                  & \textbf{2176}                \\
WaveMix-256               & 448                  & 768                 \\
\hline
\end{tabular}
\end{table}

We attribute the higher accuracy for the other architectures reported in their original papers to the the effects of various well-intentioned incremental training methods (tips and tricks), including RandAugment~\cite{cubuk2019randaugment}, mixup~\cite{zhang2017mixup}, CutMix~\cite{Yun_2019_ICCV}, random erasing~\cite{zhong2017random}, gradient norm clipping~\cite{zhang2020gradient}, learning rate warmup~\cite{Gotmare2019ACL} and cooldown, and timm augmentations~\cite{rw2019timm} . These additional methods improve the results of the core architectures by a few percentage points each. However, experimenting with these additional training methods requires extensive hyperparameter tuning. On the other hand, by excluding these methods, we were able to compare the contribution of the base architectures in a uniform manner. Even though the accuracy obtained in our experiments for the other architectures are thus slightly lower than the previously reported numbers, the results are still within the expected range when such additional training tricks are not used. Another reason for the lower performance of these models is due to small number of epochs we used for training due to resource constraints. Running the models for hundreds and thousands of epochs will give better performance.

\section{Conclusions and Discussion}

Training architectures that require large GPU RAMs for any workable batch size, whose hyperparameters have to be tuned on large clusters, have become out-of-reach for most researchers who depend on affordable GPU servers or cloud services. We re-emphasize that our objective was to propose an innovative neural architecture that can be trained on affordable hardware (e.g., a single GPU) without compromising much on classification accuracy. It was neither our objective nor within our means to pursue a singular focus on beating the state-of-the-art accuracy while disregarding the computational effort and GPU RAM required for training. 

We proposed an attention-less WaveMix architecture for token mixing for images by using 2D wavelet transform. The WaveMix architecture offers the best of both self-attention networks and CNNs by combining long distance token mixing of attention; and low GPU RAM consumption, efficiency, and speed of CNNs. It is better tailored for computer vision applications as it handles the data in 2D format without unrolling it as a sequence unlike the transformer models, such as ViT, CCT, CvT and hybrid xformers. Our experiments on image classification show that WaveMix achieves competitive accuracy with orders of magnitude lower GPU RAM consumption compared to transformer and convolutional models.

This work can be extended in several directions. Variants of the proposed architecture, such as those inspired from U-Net~\cite{UNet} and YOLO~\cite{YOLO}, will need to be tried for other computer vision tasks, such as semantic segmentation and object detection. While we tested the simplest wavelet family (Haar), other wavelet families might give better results. It can also be tested whether the wavelet family itself needs to vary with the layer depth. Alternatively, the mother wavelet  itself can be learned at different levels in an end-to-end manner. The role of the sparseness of the wavelet response at different levels can also be examined, as has been done for image compression. An additional redundancy that is not fully exploited by wavelets either is the rotational invariance, for which other mechanisms are needed.

The high accuracy of image classification by transformers and CNNs comes with high costs in terms of training data, computations, GPU RAM, hardware costs, form factors, and power consumption~\cite{li2020train}, while in several practical situations there are tight constraints on these factors. Overall, our research suggests that alternatives to convolutional or attention-based architectures for vision need to be explored to better exploit image redundancies to reduce these requirements, while still generalizing well. Neural architectures that exploit domain-specific inductive biases have previously (and in the present study) resulted in such improvements, and this search for alternative architectural innovations must continue.


\bibliographystyle{unsrtnat}
\bibliography{template}  

\end{document}